%% file: main.tex
\documentclass[10pt,twocolumn,letterpaper]{article}

\usepackage{iccv}
\usepackage{times}
\usepackage{epsfig}
\usepackage{graphicx}
\usepackage{amsmath}
\usepackage{amssymb}
\usepackage{booktabs}
\usepackage{multirow}
\usepackage{caption}
\usepackage{bbding}
\usepackage{floatrow}
\usepackage{enumitem}
\usepackage{float}

\input{def}

\usepackage[pagebackref=true,breaklinks=true,letterpaper=true,colorlinks,bookmarks=false]{hyperref}
\newcommand{\sexyname}{A$^2$Nav\xspace}

\iccvfinalcopy % *** Uncomment this line for the final submission

\ificcvfinal\fi

\begin{document}

\definecolor{MyDarkGreen}{RGB}{41, 137, 18}
\definecolor{MyBrown}{rgb}{0.76,0.35,0.06}
\definecolor{YesColor}{RGB}{0,0,0}
\definecolor{NoColor}{RGB}{0,0,0}
\def\yes{\textcolor{YesColor}}
\def\no{\textcolor{NoColor}}
\def\ph{\textcolor{blue}}
\def\todo{\textcolor{red}}
\def\sxy{\textcolor{MyDarkGreen}}

%%%%%%%%% TITLE
\title{\sexyname: Action-Aware Zero-Shot Robot Navigation by Exploiting Vision-and-Language Ability of Foundation Models}
% \title{\sexyname: Action-Aware Zero-Shot Vision-and-Language Navigation \\by Exploiting Foundation Models}

\author{
    Peihao Chen\textsuperscript{\rm 1,3} ~~ 
    Xinyu Sun\textsuperscript{\rm 1}~~ 
    Hongyan Zhi\textsuperscript{\rm 1} ~ 
    Runhao Zeng\textsuperscript{\rm 2} \\
    Thomas H. Li\textsuperscript{\rm 5} ~
    Gaowen Liu\textsuperscript{\rm 7} ~
    Mingkui Tan\textsuperscript{\rm 1,4}\thanks{Corresponding author. Email: mingkuitan@scut.edu.cn} ~
    Chuang Gan\textsuperscript{\rm 5} \\
    \textsuperscript{\scriptsize{\rm 1}}\small{South China University of Technology,}
    \textsuperscript{\scriptsize{\rm 2}}\small{Shenzhen University,}\\
    \textsuperscript{\scriptsize{\rm 3}}\small{Pazhou Laboratory,}
    \textsuperscript{\scriptsize{\rm 4}}\small{Key Laboratory of Big Data and Intelligent Robot, Ministry of Education,}\\
    \textsuperscript{\scriptsize{\rm 5}}\small{Shenzhen Graduate School, Peking University,} 
    \textsuperscript{\scriptsize{\rm 6}}\small{UMass Amherst,}
    \textsuperscript{\scriptsize{\rm 7}}\small{Cisco Research}\\
}

\maketitle
% Remove page # from the first page of camera-ready.
\ificcvfinal\thispagestyle{empty}\fi

%%%%%%%%% ABSTRACT
\begin{abstract}
We study the task of zero-shot vision-and-language navigation (ZS-VLN), a practical yet challenging problem in which an agent learns to navigate following a path described by language instructions \textbf{without requiring any path-instruction annotation data}. Normally, the instructions have complex grammatical structures and often contain various action descriptions (\eg, ``proceed beyond'', ``depart from''). How to correctly understand and execute these action demands is a critical problem, and the absence of annotated data makes it even more challenging. Note that a well-educated human being can easily understand path instructions without the need for any special training. In this paper, we propose an action-aware zero-shot VLN method (\sexyname) by exploiting the vision-and-language ability of foundation models. Specifically, the proposed method consists of an instruction parser and an action-aware navigation policy. The instruction parser utilizes the advanced reasoning ability of large language models (\eg, GPT-3) to decompose complex navigation instructions into a sequence of action-specific object navigation sub-tasks. Each sub-task requires the agent to localize the object and navigate to a specific goal position according to the associated action demand. To accomplish these sub-tasks, an action-aware navigation policy is learned from freely collected action-specific datasets that reveal distinct characteristics of each action demand. We use the learned navigation policy for executing sub-tasks sequentially to follow the navigation instruction. Extensive experiments show \sexyname achieves promising ZS-VLN performance and even surpasses the supervised learning methods on R2R-Habitat and RxR-Habitat datasets.

\end{abstract}

%%%%%%%%% BODY TEXT
\section{Introduction}

% Definition of VLN. VLN is important.
In vision-and-language navigation (VLN) tasks, an agent is required to navigate in a novel environment according to language navigation instructions. This is a crucial step towards creating intelligent agents that can interact with humans in a natural way, such as in the context of home robotics~\cite{saycan,homerobot, gan2020threedworld, gan2022threedworld} or warehouse assistants~\cite{warehouse1,warehouse2}.

\begin{figure}[t]
  \centering
    \includegraphics[width=0.9\linewidth]{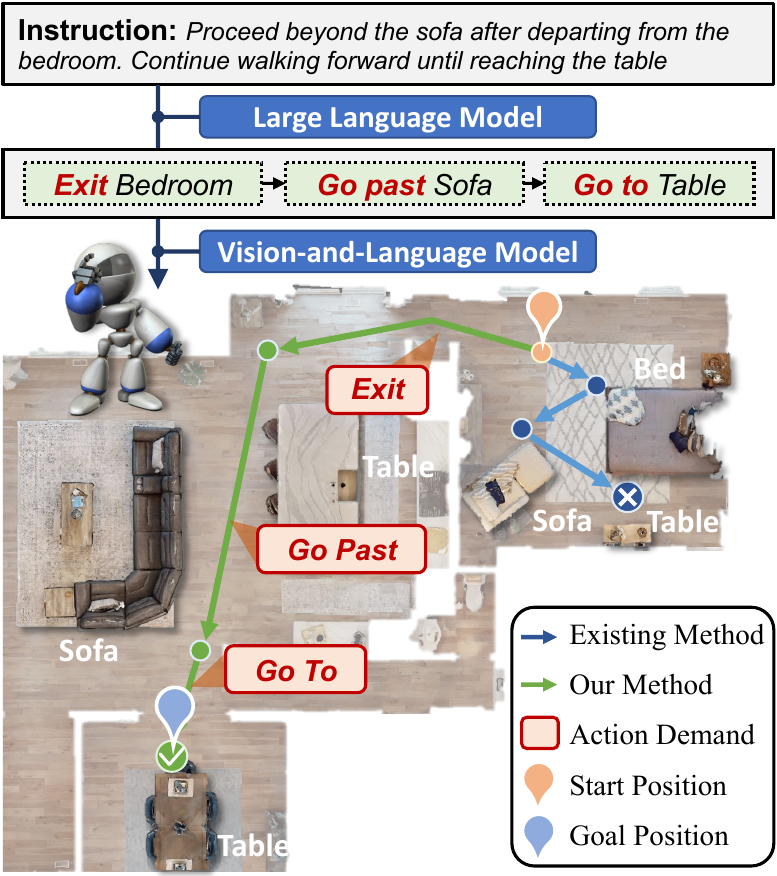}
  \caption{Existing zero-shot VLN methods navigate to the front of the landmarks sequentially, overlooking the action demands in the instruction. Our \sexyname correctly parses action demands from the instruction and accurately executes them for successfully following navigation instruction. 
  }
  \label{fig:teaser}
\end{figure}

Current dominant methods~\cite{WS-MGMap,Seq2Seq,LAW,cm2} attempt to learn VLN ability in a supervised learning manner, relying on a large amount of manually labeled path-instruction pair data. 
However, creating high-quality labeled data requires a significant amount of human effort, which can be time-consuming and expensive. Additionally, the labeled data may not cover all possible scenarios, making it challenging for the model to generalize to new, unseen environments.
To address these challenges, exploiting the knowledge from large foundation models~\cite{GPT3,CLIP,bert} for learning navigation ability without requiring downstream task annotated data is a potential solution. We call it zero-shot navigation ability.

Recently, researchers have made some attempts~\cite{cow, zson, zer, zhang2023building} at solving object navigation tasks in a zero-shot manner. They use a foundation vision-and-language model (VLM) ~\cite{CLIP} to localize the object~\cite{cow} or use it to encode the object goal features~\cite{zson}, enabling the agent to navigate to any object goal described by natural language. To utilize this zero-shot object navigation ability for the VLN task, researchers~\cite{LM-Nav,clipnav} leverage large language models (LLMs)~\cite{GPT3} to parse all landmarks in the navigation instruction, decomposing the VLN task into a sequence of object navigation sub-tasks. The agent navigates to these landmarks one by one to follow the navigation instruction.

Although some progress has been made, existing methods fail to take into account the varied action demands (\eg, ``\textit{proceed beyond}'', ``\textit{depart from}'') in instructions. This may lead the agent to the wrong destination and fail to follow the instruction. For the example in Figure~\ref{fig:teaser}, the agent is expected to ``\textit{exist the bedroom}'', but the existing methods only take the landmark ``\textit{bedroom}'' into consideration. 
As a result, the agent mistakenly goes into the bedroom, which is in the opposite direction of the path described by the instruction.
To solve this problem, the agent must correctly figure out the expected action demand associated with each landmark and accurately execute them. 
However, due to the complex grammatical structure and diverse action expressions in the instruction, correctly understanding the action demands is non-trivial. More critically, under the condition that no path-instruction annotation is available, how to learn a navigation policy that is able to execute these action demands is still an open problem.

In this paper, we propose an action-aware navigation method, named \sexyname, for the zero-shot VLN task. Our method consists of two components: an instruction parser for figuring out landmarks and associated action demands; and an action-aware navigation policy for executing these action demands sequentially for navigation.

Specifically, we leverage the reasoning ability of LLMs for decomposing an instruction into a sequence of action-specific object navigation sub-tasks. Each sub-task (\eg, ``\textit{go past sofa}'') requires the agent to localize the landmark and navigate to an expected goal position.  
According to the action demands that lead the agent to different goal positions, we summarize five fundamental sub-tasks and learn an action-specific object navigator for each of them. 
However, due to the absence of labeled data for training these navigators, we adopt the approach proposed in ZSON~\cite{zson} to transform the task of training an object navigator into training an image-goal navigator using a freely collected dataset.
In detail, we first encode both the text description and image of the landmark to a shared  semantic feature space using CLIP~\cite{CLIP}. 
Then we construct a dataset by randomly sampling a path and capturing an image of the landmark at a specific location based on the corresponding action demand.
For instance, for the ``\textit{go past}'' action demand, the landmark is usually located in the middle of the navigation path. The image-goal navigator trained on this dataset could be used for action-specific object navigation given a CLIP-encoded textual landmark.

Extensive experiments demonstrate that our \sexyname achieves promising performance on zero-shot VLN task, getting 22.6\% and 16.8\% success rates on R2R-Habitat and RxR-Habitat, respectively without requiring any path-instruction annotation. Notably, the proposed \sexyname even outperforms the current state-of-the-art supervised learning method~\cite{WS-MGMap} by 1.8\%  on RxR-Habitat.
In summary, our main contributions are as follows:
\begin{itemize}[leftmargin=3mm]
    \item Instead of treating vision-and-language navigation as a sequence of object navigation tasks, we take into account the instruction action demands and decompose the instruction into a sequence of action-specific object navigation sub-tasks, where the agent is expected to not only localize the landmarks but also navigate to different goal position according to the associated action demand. 
    \item To address the problem that existing zero-shot navigators cannot satisfy different action demands, we identify and summarize five fundamental action demands and learn a unique navigator for executing each one without requiring manual path-instruction annotation, leading to more accurate and explainable navigation results.
\end{itemize}

\begin{figure*}[t]
  \centering
    \includegraphics[width=1\linewidth]{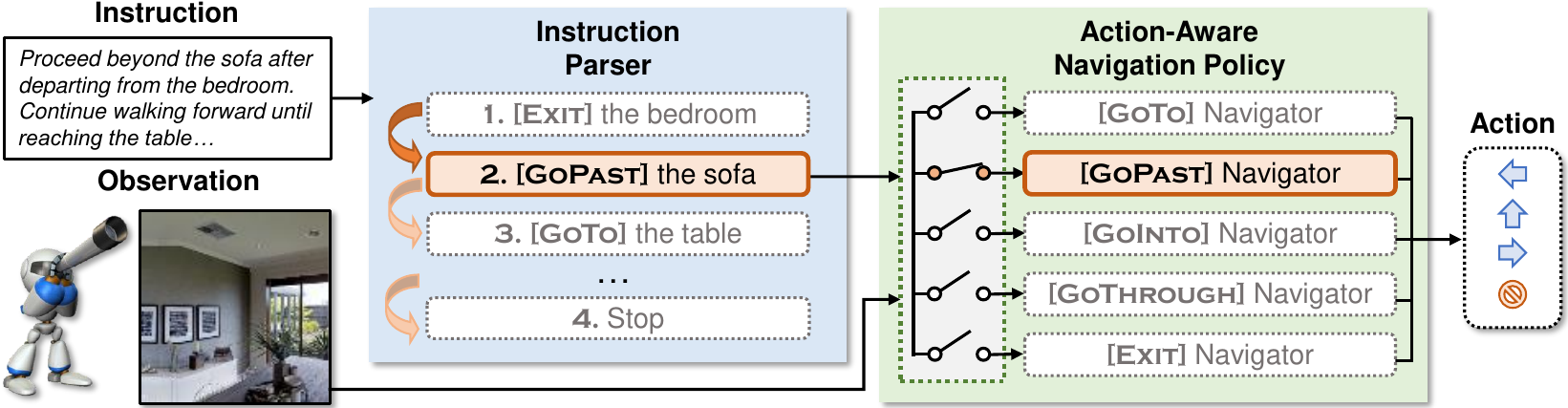}
  \caption{General scheme of \sexyname for zero-shot VLN task. \sexyname consists of an instruction parser for decomposing an instruction into action-specific object navigation sub-task sequence, and an action-aware navigation policy for executing these sub-tasks sequentially.}
    % \vspace{-2mm}

  \label{fig:overview}
\end{figure*}

\section{Related Works}

\subsection{Vision-and-Language Navigation}
Existing VLN methods can be summarized into two categories. The first category of methods leverage reinforcement learning method~\cite{vgni,lbyl,rcmm}, data augmentation~\cite{Speaker-Follower,envdrop}, auxiliary tasks~\cite{smna,ssar,rcma} and pre-training~\cite{DUET,VLN-BERT,tlag} to perform navigation on discrete environments where only points in a sparse set are navigable. We focus on another category of VLN methods that perform navigation on continuous environments~\cite{btng,WS-MGMap,cm2,ccc,wmfl,WPN,tpwt,vlsf,ding2022embodied}. Among these methods, Krantz \etal~\cite{btng} are the first to propose a continuous VLN setting based on the Habitat simulator~\cite{habitat}. Qi \etal~\cite{kwkw} further propose to train the agent in an end-to-end manner that takes step observation and instruction as input to predict an action for each step. Other methods like waypoint prediction~\cite{wmfl} and semantic top-down map construction~\cite{WS-MGMap,cm2} are proposed to enhance the navigation performance.
However, these methods rely heavily on manual-labeled path-instruction pairs to train the agent in a fully supervised manner, thus are not scalable to a new scenario without any path-instruction pair data.

\subsection{Zero-Shot Object Navigation}
Since the navigation instruction is often described by several landmarks, it can be decomposed into sequential object navigation tasks. The object navigation task
% the agent is asked to find a specific object and correctly navigate to the front of it, which 
has been explored by previous literature~\cite{cow,zson,etess,acmon,PONI,zer,gose,OVRL}. Among these methods, we notice that some trails that design the object navigation agent in a zero-shot manner show great potential. Gadre \etal~\cite{cow} design a heuristic algorithm to navigate to an object using the open-world object recognition ability of the foundation vision-and-language model (\ie, CLIP~\cite{CLIP}). Some works like ZER~\cite{zer} and ZSON~\cite{zson} learn an image navigation agent first, and then map the image goal representation into object text goal embedding space, and thus transfer to the object navigation task. In this paper, we follow ZSON~\cite{zson} to use a pre-trained foundation model CLIP to map the goal representation into joint vision-and-language embedding space and thus train an object navigator using randomly collected image-path pairs.

\subsection{Zero-Shot Vision-and-Language Navigation}
Based on the previous success on VLN and zero-shot object navigation, we aim to tackle the VLN task in the zero-shot manner, releasing the agent from expensive manual-labeled path-instruction training data. This problem has not been fully exploited yet. Pioneering works~\cite{LM-Nav,clipnav,chen2023see} have already verified the effectiveness of foundation models (LLM~\cite{GPT3} and VLM~\cite{CLIP}) in this scenario. These methods leverage GPT-3~\cite{GPT3} to extract navigation landmarks from the instruction and then initialize a heuristic object navigator using CLIP~\cite{CLIP} to find out the landmark from visual observation and to navigate to the front of the landmark. 
However, these methods neglect actions in the instructions since they can only directly go to the landmark. 
Concurrent work~\cite{navgpt} leverages a GPT model for inferring navigation actions on a discrete navigation graph. The performance in continuous environments has not been well explored. 
Our proposed  \sexyname solved these issues using a learnable action-aware object navigator.

\section{Action-Aware Zero-Shot VLN}

\subsection{Problem Definition}
In the vision-and-language navigation (VLN) task, an agent performs a series of low-level actions (\eg, \textsc{GoForward}, \textsc{TurnLeft} and \textsc{TurnRight}) in a novel continuous environment to follow a specific path $P$ described by the instruction $I$. We consider a more practical but challenging problem zero-shot VLN, where the agent is expected to complete the VLN task without requiring path-instruction annotation. Existing methods attempt to address this challenging problem by parsing all landmarks in the instruction and decomposing the VLN task into a series of object navigation sub-tasks. Then the agent navigates to these landmarks sequentially. However, these methods overlook the specific action demands associated with each landmark (\eg, ``\textit{go to}'', ``\textit{go past}'' ), which could potentially lead to the agent failing to follow the instructions correctly.

To solve this problem, we leverage a large language model as an instruction parser for parsing all landmarks and their associated action demands. The instruction is then decomposed into a sequence of action-specific object navigation sub-tasks, in which the agent is required to localize the landmark and navigate based on the specific action demands associated with that landmark. For executing each sub-tasks sequentially, an action-aware navigation policy comprising five action-specific navigators is learned in a zero-shot manner. The general scheme is shown in Figure~\ref{fig:overview}.

\subsection{Instruction Parser}
\label{sec:instruction_parser}
The instruction parser aims to transfer a complicated linguistic instruction into several sequential executable action-specific object navigation sub-tasks~\cite{fgr2r,chen2021grounding,chencomphy}.
We first introduce the definition of action-specific object navigation sub-tasks, followed by a description of how we use LLMs to decompose an instruction into these sub-tasks.

\subsubsection{Action-Specific Object Navigation Sub-Task}
Each action-specific object navigation sub-task contains a landmark and an associated action demand, such as ``\textit{departing from the bedroom}''. The sub-task can be represented by a template ``(\textsc{Action}, \textsc{Landmark})''. 

We empirically observe that for the same landmark, different action demands result in different spatial relationships between the landmark and navigation path.  The landmark can be located at the beginning, middle, or end of a navigation path.
For example, the landmark is located at the end of the path for action demand ``\textit{go to the landmark}'', while located in the middle for ``\textit{go past the landmark}''.
Besides, the landmark could be an object or a region. 

\begin{figure}[t]
  \centering
    \includegraphics[width=1\linewidth]{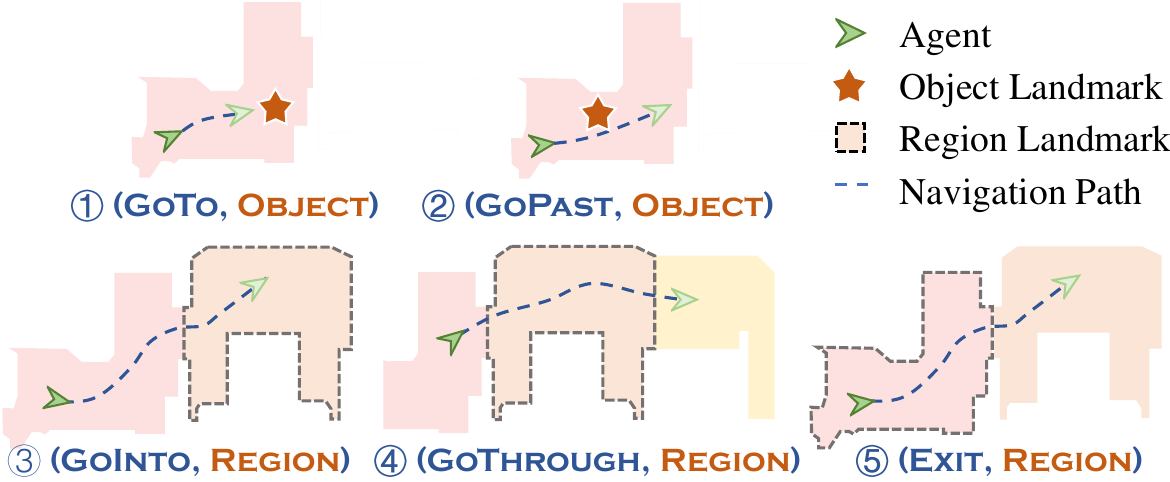}
  \caption{Visualization of different sub-task types. 
  % The landmarks can be an object or a region. 
  For different action demands, the landmark is located at a different position related to the path.
  }
  \label{fig:dataset}
  % \vspace{-6mm}
\end{figure}

Based on these two observations, we summarize the following basic sub-tasks, namely, ``{(\textsc{GoTo}, \textsc{Object})}'', ``{(\textsc{GoPast}, \textsc{Object})}'', ``{(\textsc{GoAway}, \textsc{Object})}'', and ``{(\textsc{GoInto}, \textsc{Region})}'', ``{(\textsc{GoThrough}, \textsc{Region})}'', ``{(\textsc{Exit}, \textsc{Region})}''.
Since the ``(\textsc{GoAway}, \textsc{Object})'' sub-task does not indicate a specific navigation destination, it can cause confusion for the agent during executing. Therefore, we remove it from the basic sub-task list.
An illustrative diagram for these sub-tasks is shown in Figure~\ref{fig:dataset}.

\subsubsection{Decomposing Instruction into Sub-Tasks} 
We use the GPT-3 LLM~\cite{GPT3} for decomposing an instruction into a sequence of sub-task described above. We use a prompt with several correct instruction parsing examples for telling GPT-3 the parsing requirements. We have tried different prompt designs and found that this prompt yields the best performance (cf. Section~\ref{sec:ablation_instruction_parser}). More details about the prompt design can be referred to Appendix.

As the predicted sub-tasks from GPT-3 are in the free-form language, we need to map each prediction to the predefined sub-tasks. In most cases, the ``\textsc{Action}'' predictions made by GPT-3 accurately match one of the ``\textsc{Action}'' in predefined sub-tasks, and thus we can directly map it to this sub-task type. In cases where a prediction does not match any, we follow~\cite{planner} to perform mapping through semantic translation. Specifically, we use BERT~\cite{bert} to encode the predicted ``\textsc{Action}'' and the ``\textsc{Action}'' in all predefined sub-tasks. Then we compute the cosine similarity between them and consider the predefined sub-task with the highest score as the predicted sub-task.

\begin{figure}
    \centering
    \includegraphics[width=1\textwidth]{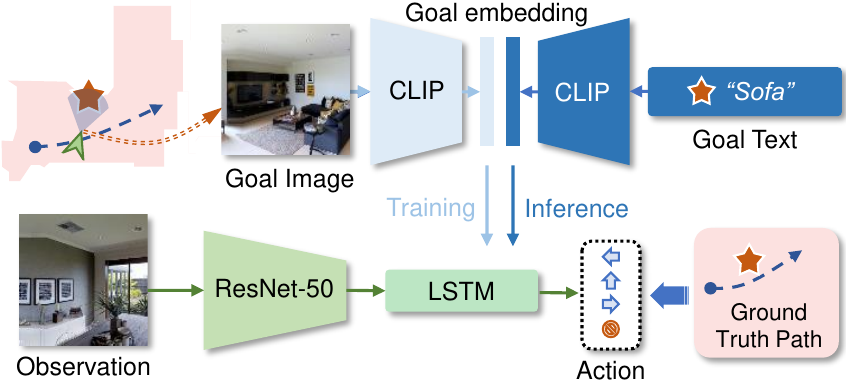}
    \caption{Training and inference pipeline of action-specific object navigators.}
    \label{fig:navigator}
    % \vspace{-4mm}
\end{figure}

\subsection{Action-Aware Navigation Policy}

With the sub-task sequence parsed by LLMs, we learn an action-aware navigation policy to execute them sequentially using low-level navigation actions.
The policy consists of five action-specific navigators, each of which is responsible for a specific sub-task type. 
We follow ZSON~\cite{zson} to transform the question of learning such a navigator into learning an image-goal navigator on a freely collected action-specific image-path dataset.
In this section, we first review ZSON, followed by a description of how we collect such a dataset and learn action-specific navigators on this dataset.

\subsubsection{Revisiting Zero-Shot Object Navigator ZSON}
ZSON~\cite{zson} is a zero-shot object navigator that enables an agent to navigate to a landmark without the need for any annotations.
It achieves this ability through two steps. First, it generates a dataset of image-path pairs by randomly sampling navigation paths in different environments and capturing a goal image at the end of each path. 
Second, it encodes the goal image to a semantic feature space using CLIP~\cite{CLIP}. 
An image-goal navigator is trained on this dataset using a reinforcement learning algorithm (\ie, DD-PPO~\cite{ddppo}) for navigating to a goal position where the goal image is taken.
The trained image-goal navigator can be used for object navigation by encoding the object text to the same semantic feature space using CLIP.
Since collecting the image-path pair data does not need any object annotation or path annotation, ZSON can be trained in a zero-shot manner. 

Despite its effectiveness in the object navigation task, it overlooks the action demands and can only navigate to the front of the landmark. This cannot meet the requirement of action-specific object navigation sub-tasks and may lead to the failure of the VLN task.

\subsubsection{Learning Action-Specific Object Navigator}
We fine-tune the trained ZSON model on five action-specific image-path datasets for learning five unique action-specific navigators, respectively. 
In each action-specific dataset, the location of landmarks and paths should reveal the characteristic of the corresponding action demand.
Next, we introduce the principle for collecting these datasets. By default, the path is randomly sampled and the image is captured at the end of the path.

For the \textbf{\textsc{GoTo}} action demands, we directly utilize a trained ZSON model as a navigator without fine-tuning since this action demand can be well handled by ZSON.
For the \textbf{\textsc{GoPast}} action demand that expects the agent to go to the object and keep going forward past the object, we capture the goal image in the middle of the path. 
For the \textbf{\textsc{GoInto}} action demand that expects the agent to go cross a doorway into the target region, we sample the path that crosses over two regions and sample the goal image at the end of this path.
For the \textbf{\textsc{GoThrough}} action demand that expects the agent to go from one side to the other side of a region, we randomly sample the path that starts near one entrance and ends near the other one of a region. The goal image is captured in the middle of the path. 
For the \textbf{\textsc{Exit}} action demand, the path is sampled in the same way as the \textsc{GoInto} action demand, while the goal image is captured at the beginning of the path.

We fine-tune the trained ZSON model on these datasets using the same learning pipeline as ZSON, which is shown in Figure~\ref{fig:navigator}. 
More details can be found in Appendix.

\begin{table*}[ht]
\renewcommand\arraystretch{1.0}
\centering
\small
\begin{tabular}{@{}clcccccc@{}}
\toprule
\multicolumn{1}{l}{\multirow{2}{*}{}} & \multirow{2}{*}{Method}    & \multirow{2}{*}{Extra Info.} & \multicolumn{2}{c}{R2R-Habitat}           & \multicolumn{2}{c}{RxR-Habitat}          & \multirow{2}{*}{CSR} \\ \cmidrule(lr){4-5} \cmidrule(lr){6-7}
\multicolumn{1}{l}{}                  &                            &                              & SR              & SPL             & SR              & SPL            &                      \\ \midrule
\multirow{3}{*}{Supervised}           & Seq2Seq~\cite{Seq2Seq}     & Depth                        & 25.0\%          & 22.0\%          & -               & -              & -                    \\
                                      & LAW~\cite{LAW}             & Depth                        & 35.0\%          & 31.0\%          & 10.0\%          & 9.0\%          & 28.6\%               \\
                                      & WS-MGMap~\cite{WS-MGMap}   & Depth                        & 38.9\%          & 34.3\%          & 15.0\%          & 12.1\%         & 38.6\%               \\ \midrule
\multirow{6}{*}{Zero-Shot}            & Random                     & -                            & 0.0\%           & 0.0\%           & 6.0\%           & 6.0\%          & 0.0\%                \\
                                      & CLIP-Nav~\cite{clipnav}    & Panoramic                    & 5.6\%           & 2.9\%           & 9.8\%           & 3.2\%          & 57.4\%               \\
                                      & Seq CLIP-Nav~\cite{clipnav}& Panoramic                    & 7.1\%           & 3.7\%           & 9.1\%           & 3.3\%          & 77.8\%               \\
                                      & Cow~\cite{cow}             & Depth                        & 7.8\%           & 5.8\%           & 7.9\%           & 6.1\%          & \textbf{98.3\%}      \\
                                      & ZSON~\cite{zson}           & -                            & 19.3\%          & 9.3\%           & 14.2\%          & 4.8\%          & 73.6\%               \\ \cmidrule(l){2-8} 
                                      & \textbf{\sexyname (Ours)}  & -                            & \textbf{22.6\%} & \textbf{11.1\%} & \textbf{16.8\%} & \textbf{6.3\%} & 74.3\%               \\ \bottomrule
\end{tabular}
\caption{Comparisons with zero-shot and supervised methods on VLN datasets. The depth and panoramic in ``Extra Info.'' column indicate that these methods depend on an extra depth sensor or panoramic sensor.}
\label{tab:sota}
\end{table*}

\subsection{Zero-Shot Vision-and-Language Navigation}
We use the learned action-aware navigation policy for executing GPT-3 predicted sub-tasks sequentially. Specifically, we first identify the current sub-task type and select the corresponding action-specific navigator for predicting a low-level action. 
When the navigator predicts a \textsc{STOP} action or exceeds the sub-task maximum step $m_s$, we switch to the next sub-task. This process repeats until all sub-tasks are finished or the episode maximum step $m_e$ is reached.

\section{Experiments}
\subsection{Experimental Setup}

\paragraph{Evaluation Datasets and Metrics.} We conduct experiments on the validation unseen split of three datasets, namely R2R-Habitat~\cite{vlnce}, RxR-Habitat~\cite{vlnce}, and Fine-Grained R2R (FG-R2R)~\cite{fgr2r}. These three datasets contain 1,839, 1,079, and 1,839 validation episodes on 11 scenes in Matterport3D, respectively. 
RxR-Habitat contains instructions in three languages, and we only use the English split in our experiments.
FG-R2R is an extension of R2R~\cite{r2r}, where instructions are chunked into several sub-instructions and each sub-instruction is labeled with a corresponding sub-path, resulting in 6,687 sub-instruction-sub-path pairs.
Following the existing works~\cite{Seq2Seq,LAW,WS-MGMap}, we evaluate VLN performance using Success Rate (\textbf{SR}) and Success weighted by inverse Path Length (\textbf{SPL}).
Besides, to evaluate the generalization ability among datasets, we follow Dorbala \etal~\cite{clipnav} to propose Consistency on SR (\textbf{CSR}) for computing the relative change in SR among datasets. Specifically, $ \rm{CSR} = 1 - \frac{|SR_a - SR_b|}{\max\{SR_a,~SR_b\}} \times 100\%$, where $\rm SR_{a}$ and $\rm SR_{b}$ are success rates for different datasets.

\paragraph{Agent Configurations.}
Following ZSON~\cite{zson}, the agent has a height of 1.25m, with a radius of 0.1m. It is equipped with one 128 $\times$ 128 RGB sensor with 90$^{\circ}$ horizontal field of view. The agent can execute four low-level actions, namely \textsc{STOP} indicating the end of an episode, \textsc{Forward} that moves itself forward by 0.25 meters and \textsc{TurnLeft} and \textsc{TurnRight} that turn itself by 30$^{\circ}$. The $m_s$ is empirically set to 100 and 50 for R2R-Habitat and RxR-Habitat, respectively. The $m_e$ is set to 500 for all three datasets following exiting works~\cite{WS-MGMap,LAW}.
\vspace{-3mm}

\paragraph{Baselines.} 
We decompose the instruction into an object navigation sub-task sequence using GPT-3 and execute these sub-tasks sequentially using four zero-shot object navigation methods.

\begin{itemize}[leftmargin=3mm]
\vspace{-3mm}
    \item \textbf{CLIP-Nav}~\cite{clipnav} is designed for navigating among discrete navigable nodes. The agent uses CLIP to determine which adjacent node has the highest possibility of containing landmarks and then moves to this node. We adapt it to continuous environments using a waypoint navigation algorithm. More details can be found in Appendix.
\vspace{-7mm}

\item \textbf{Seq CLIP-Nav}~\cite{clipnav} is an extended version of CLIP-Nav with an additional backtracking mechanism, which allows the agent to go back to the previous location if it cannot find the landmarks for several steps.
\vspace{-2mm}

\item \textbf{CoW}~\cite{cow} uses CLIP gradient for object localization and a path-planning algorithm for action determination. 
\vspace{-2mm}

\item \textbf{ZSON}~\cite{zson} uses the CLIP for encoding both image and landmark text to the same semantic feature space and then trains an image navigator for object navigation.
% \vspace{-3mm}
\end{itemize}

\subsection{Zero-Shot Vision-and-Language Navigation }
We compare our \sexyname with existing zero-shot navigation methods on R2R-Habitat and RxR-Habitat datasets. Besides, we compare our method with supervised learning methods and study whether our zero-shot method is able to outperform supervised learning methods when the number of annotated data is limited.

\subsubsection{Comparisons with Zero-Shot Methods}
In Table~\ref{tab:sota}, leveraging random actions failed to reach the goal position in all episodes on R2R-Habitat dataset, which demonstrates that zero-shot VLN is a challenging problem. On RxR-Habitat, random walking gets a 6\% success rate because almost 6\% of episodes have a goal position within the success distance of 3 meters. These episodes are  judged to be successful even if the agent simply remains stationary. Compared with random walking, all other zero-shot navigation methods have a significant improvement, while our \sexyname performs the best on two benchmark datasets, outperforming CLIP-Nav, Seq CLIP-Nav, CoW and ZSON by 17.0\%, 15.5\%, 14.8\%, and 3.3\%, respectively on R2R-Habitat, and by 7.0\%, 7.7\%, 8.9\%, and 2.6\%, respectively on RxR-Habitat in terms of success rate. 
We provide a more detailed analysis of these zero-shot methods below.

Existing zero-shot navigation methods can be categorized into two groups: 1) using CLIP to localize the landmark and then using a path-planning algorithm for navigation (\ie, CoW, CLIP-Nav, and Seq CLIP-Nav); 2) using CLIP to encode the landmark and then learning a policy for navigation (\ie, ZSON). In Table~\ref{tab:sota}, the former approaches show inferior performance. They struggle to determine the exploration direction when the landmark is out of the field of view. We hypothesize this is due to the difficulty of CLIP in utilizing room layout to infer the possible location of landmarks. 
ZSON eases this problem by training a navigation policy that incorporates room layout commonsense to identify the most likely area to find the landmark.
Our \sexyname surpasses these methods by incorporating action-aware navigation skills.

\begin{figure}[t]
  \centering
    \includegraphics[width=1\linewidth]{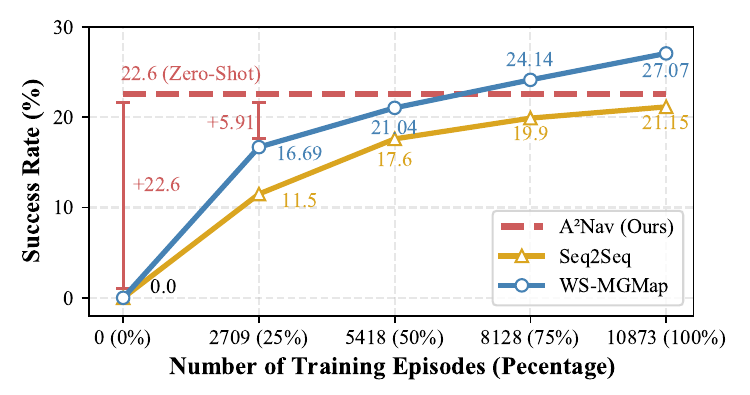}
  \caption{Comparison with the supervised learning methods that are trained on partial training data. }
  \label{fig:vis}
  % \vspace{-4mm}
\end{figure}

\begin{table*}[t]
\renewcommand\arraystretch{1.0}
\centering
\small
% \resizebox{}{!}{
\begin{tabular}{@{}ccccccccc@{}}
\toprule
\multirow{2}{*}{GO TO} & \multirow{2}{*}{GO PAST} & \multirow{2}{*}{GO INTO} & \multirow{2}{*}{GO THROUGH} & \multirow{2}{*}{EXIT} & \multicolumn{2}{c}{R2R-Habitat}           & \multicolumn{2}{c}{RxR-Habitat}          \\ \cmidrule(l){6-7} \cmidrule(l){8-9} 
                       &                          &                          &                             &                       & SR              & SPL             & SR              & SPL            \\ \midrule
\yes{\Checkmark}       & \no{\XSolidBrush}        & \no{\XSolidBrush}        & \no{\XSolidBrush}           & \no{\XSolidBrush}     & 19.3\%          & 9.3\%           & 14.2\%          & 4.8\%          \\
\yes{\Checkmark}       & \yes{\Checkmark}         & \no{\XSolidBrush}        & \no{\XSolidBrush}           & \no{\XSolidBrush}     & 20.9\%          & 9.9\%           & 14.8\%          & 5.1\%          \\
\yes{\Checkmark}       & \yes{\Checkmark}         & \yes{\Checkmark}         & \no{\XSolidBrush}           & \no{\XSolidBrush}     & 21.5\%          & 9.8\%           & 14.7\%          & 5.2\%          \\
\yes{\Checkmark}       & \yes{\Checkmark}         & \yes{\Checkmark}         & \yes{\Checkmark}            & \no{\XSolidBrush}     & 22.3\%          & 10.7\%          & 16.7\%          & 6.1\%          \\
\yes{\Checkmark}       & \yes{\Checkmark}         & \yes{\Checkmark}         & \yes{\Checkmark}            & \yes{\Checkmark}      & \textbf{22.6\%} & \textbf{11.1\%} & \textbf{16.8\%} & \textbf{6.3\%} \\ \bottomrule
\end{tabular}
% }
\caption{Ablation study on action-aware navigation policy with different combinations of action-specific navigators.}
\label{tab:ablation-vln}
\end{table*}

% Please add the following required packages to your document preamble:
% \usepackage{multirow}
\begin{table*}[t]
\renewcommand\arraystretch{1.0}
\centering
\small
% \resizebox{0.8\linewidth}{!}
% {
\begin{tabular}{@{}lcccccccccc@{}}
\toprule
\multirow{2}{*}{Methods} & \multicolumn{2}{c}{GO TO}         & \multicolumn{2}{c}{GO PAST}       & \multicolumn{2}{c}{GO INTO}       & \multicolumn{2}{c}{GO THROUGH}    & \multicolumn{2}{c}{EXIT}          \\ \cmidrule(l){2-3} \cmidrule(l){4-5} \cmidrule(l){6-7} \cmidrule(l){8-9} \cmidrule(l){10-11} 
                         & SR              & SPL             & SR              & SPL             & SR              & SPL             & SR              & SPL             & SR              & SPL             \\ \midrule
Random                   & 4.3\%           & 4.3\%           & 0.9\%           & 0.9\%           & 4.1\%           & 4.1\%           & 2.6\%           & 2.6\%           & 4.4\%           & 4.4\%           \\
CLIP-Nav~\cite{clipnav}                 & 16.7\%          & 12.2\%          & 11.4\%          & 7.4\%           & 15.1\%          & 10.2\%          & 13.8\%          & 9.7\%           & 15.3\%          & 9.2\%           \\
Seq CLIP-Nav~\cite{clipnav}             & \textbf{18.3\%} & \textbf{12.7\%} & 10.3\%          & 6.8\%           & 15.3\%          & 9.6\%           & 12.8\%          & 8.6\%           & 16.1\%          & 9.6\%           \\
CoW~\cite{cow}                      & 8.8\%           & 8.1\%           & 9.0\%           & 8.4\%           & 7.5\%           & 6.5\%           & 7.8\%           & 7.0\%           & 19.2\%          & 16.1\%          \\
ZSON~\cite{zson}                     & 13.8\%          & 8.2\%           & 10.0\%          & 6.3\%           & 13.2\%          & 6.9\%           & 13.5\%          & 8.1\%           & 13.0\%          & 7.0\%           \\ \midrule
\textbf{\sexyname (Ours)}& 13.8\%          & 8.2\%           & \textbf{15.0\%} & \textbf{11.7\%} & \textbf{18.4\%} & \textbf{12.1\%} & \textbf{16.1\%} & \textbf{11.0\%} & \textbf{41.1\%} & \textbf{36.9\%} \\ \bottomrule
\end{tabular}
% }
\caption{Comparisons of different zero-shot navigators on Fine-Grained R2R datasets.}
\label{tab:ablation-fgr2r}
\end{table*}

% \vspace{-0.5cm}
\subsubsection{Comparisons with Supervised Learning Methods}
We compare our method with three supervised VLN methods, namely a vanilla Seq2Seq~\cite{vlnce} which is often considered as a baseline method, LAW~\cite{LAW} and WS-MGMap~\cite{WS-MGMap} which are current state-of-the-art methods. It is worth noting that all these supervised methods take depth images and larger 224 $\times$ 224 RGB images as input and require about 10K annotated episodes for supervised training.

On R2R-Habitat, our zero-shot \sexyname achieves comparable performance in terms of SR (22.6\% \vs 25.0\%) compared with the vanilla Seq2Seq in Table~\ref{tab:sota}. 
Upon comparison with the state-of-the-art supervised method WS-MGMap, which utilizes a multi-granularity map for environment representation, our zero-shot approach exhibits a relatively large performance gap. However, we believe that incorporating a map representation into our action-aware navigation policy may enhance its performance and narrow this gap. we leave this as a direction for future work

On RxR-Habitat, \sexyname outperforms all supervised learning methods in terms of SR, increasing it from 15.0\% to 16.8\%. Besides, our method achieves a higher Consistency on the SR score, indicating that \sexyname is more effective at generalizing to different datasets and can adapt more easily to varying environments. This is a crucial advantage in real-world applications where the agent must navigate through diverse and constantly changing environments.

% \todo{[TODO: comparison on limited training data]}
We also compare our \sexyname with supervised methods that trained on partial s. For fair comparison, we re-implement two supervised VLN methods Seq2Seq~\cite{Seq2Seq} and WS-MGMap~\cite{WS-MGMap} without augmented training data and taking input 128$\times$128 image, following our default setting. In Figure~\ref{fig:vis}, our \sexyname outperforms the Seq2Seq even using all training episodes on R2R-Habitat. Compared with the state-of-the-art method with a semantic top-down map as input, our \sexyname still yields better results if less than 50\% training episodes are available for it.

% To study the performance of supervised learning methods with limited annotated data, we randomly sample a part of annotated path-instruction pairs for training WS-MGMap. In Figure~\ref{}, our \sexyname outperforms the state-of-the-art method when it uses 25\% annotated data (2,704 episodes) for training. This demonstrates the effectiveness of \sexyname under the sparse annotation condition.

\begin{figure*}[t]
  \centering
    \includegraphics[width=1\linewidth]{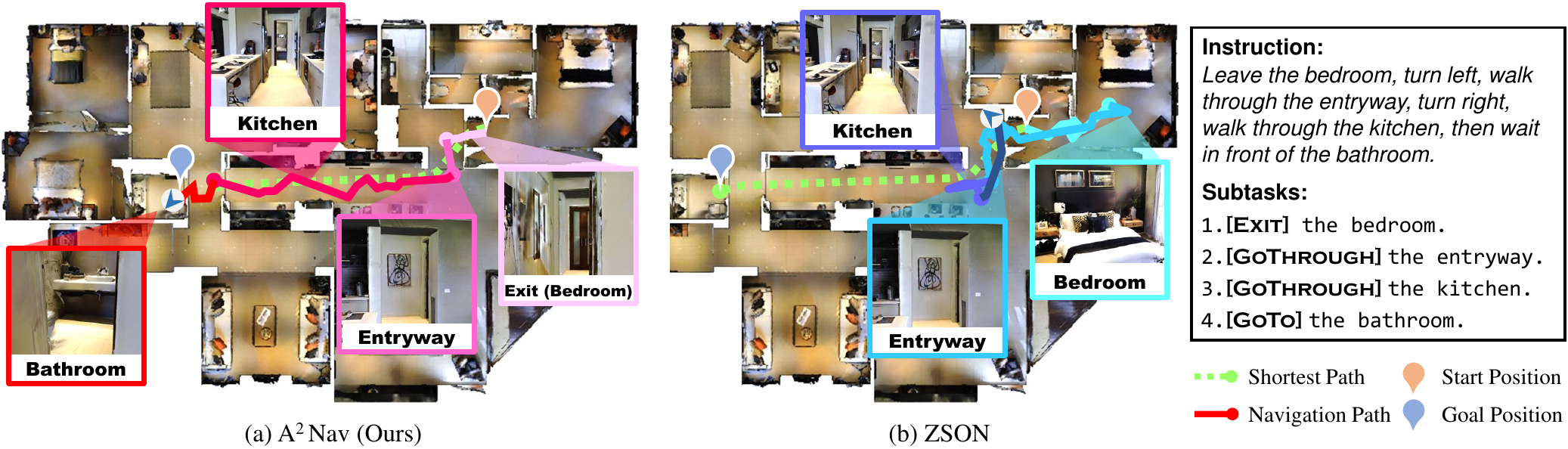}
  \caption{Qualitative example of (a) successful navigation using our \sexyname and (b) corresponding navigation result using ZSON. Paths with different colors (from light to dark) belong to sequential sub-tasks.}
  \label{fig:example-1}
  % \vspace{-6mm}
\end{figure*}

\subsection{Ablation Studies}

\subsubsection{Effectiveness of Action-Aware Navigation Policy}
To verify the effectiveness of each navigator of our action-aware navigation policy, we create multiple navigation policy variants with different combinations of navigators and evaluate them on the VLN task. By default, the sub-task is executed by the \textsc{GoTo} navigator if its corresponding navigator is not included. 
In Table~\ref{tab:ablation-vln}, the navigation policy that incorporates more sub-task navigators consistently achieves better results on both the R2R-Habitat and RxR-Habitat datasets in terms of SR and SPL.
We attribute this improvement to the fact that the sub-task navigators aid the agent in comprehending the action requirements and navigating to the target location of the current sub-task. Consequently, the agent locates the next landmark more efficiently, leading to an increased success rate in navigation.

We also evaluate its performance on different types of sub-task. To create an evaluation dataset, we parse the sub-instruction in FG-R2R into a ``(\textsc{Action}, \textsc{Landmark})'' sub-task using the instruction parser in Section~\ref{sec:instruction_parser}, and consider the sub-path as ground truth path for this sub-task. 
In Table~\ref{tab:ablation-fgr2r}, our navigation policy outperforms other zero-shot object navigation methods for 4 out of 5 sub-task types. Notably, for the ``(\textsc{Exit}, \textsc{Region})'' sub-task, \sexyname outperforms all other navigators by a large margin, increasing the SR from 19.2\% to 41.1\%. We attribute this improvement to the fact that the exiting navigators can only move  toward the landmarks, which is opposite to the expected navigation direction.
As for ``(\textsc{GoTo}, \textsc{Object})'' sub-task, we directly use the trained ZSON and thus we achieve the same performance as ZSON.

\begin{table}[t]
\begin{tabular}{cccc}
\toprule
\begin{tabular}[c]{@{}c@{}}Instruction \\ Parser\end{tabular} & Prompt                                    & SR              & SPL             \\ \midrule
Heuristic                                                          & -                                         & 17.1\%          & 9.1\%           \\ \midrule
GPT-3                                                          & Sub-Task Definition                       & 8.6\%           & 3.8\%           \\
GPT-3                                                          & Parsing Examples                          & \textbf{22.6\%} & \textbf{11.1\%} \\
GPT-3                                                          & \multicolumn{1}{l}{Definition + Examples} & 19.0\%          & 8.5\%           \\ 
\bottomrule
\end{tabular}
\caption{Comparisons of different instruction parsers. The GPT-3 with appropriate prompts performs the best.}
\label{tab:ablation-parser}
% \vspace{-2mm}
\end{table}

\vspace{-0.5mm}
\subsubsection{Comparisons on Different Instruction Parser}
\label{sec:ablation_instruction_parser}
The instruction parser aims to decompose the natural human instruction into several action-specific object navigation sub-tasks. In this paper, we use a GPT-3 LLM as the instruction parser due to its advanced reasoning ability. We also implement a heuristic variant that parses the instruction based on grammatical rules. Specifically, we follow FG-R2R~\cite{fgr2r} to chunk the instruction into several parts based on the grammatical relations between words. Each part is an independent navigation task. The verbs and the remaining words in a part are considered the action and landmark of a sub-task, respectively. We then map these free-form actions to one of five action-specific object navigation sub-task using semantic translation~\cite{planner}. In Table~\ref{tab:ablation-parser}, GPT-3 with appropriate prompts significantly outperforms the heuristic variant on R2R-Habitat dataset. We attribute this to the in-depth reasoning ability of GPT-3, which helps to figure out the correct action demands and temporal nature of landmarks from complicated long instructions.

We also evaluate the effect of different prompts for GPT-3. To tell the GPT-3 about the instruction parsing requirement, we have tried three prompt designs: a brief description of the sub-task definition, a collection of instruction parsing examples, and a combination of both.
In Table~\ref{tab:ablation-parser}, with several correct parsing examples as prompt, GPT-3 performs the best. We attribute this to its in-context learning ability~\cite{incontext}. We also observe that providing a brief sub-task description decreases performance. We hypothesize this is because such descriptions are often not detailed enough to cover all possible scenarios, thereby potentially misleading GPT-3 in its understanding of each sub-task

\vspace{-0.5mm}
\subsection{Visualization Results}
In Figure~\ref{fig:example-1}, we provide qualitative visualization results on R2R-Habitat. We visualize the trajectory of each sub-tasks with different colors. Figure~\ref{fig:example-1} (a) shows successful navigation to a bathroom according to the instruction. Based on our action-aware navigation policy, the agent finds the exit of the bedroom to go out, and then goes through the entryway and the kitchen, straightly going to the bathroom. In contrast, in Figure~\ref{fig:example-1} (b), another agent with ZSON object navigation policy directly goes to the bed that satisfies the ``bedroom" landmark in the beginning, leading to an entirely opposite direction. Subsequently, the ZSON agent goes to the entryway and stops at the near end of the kitchen, and then struggles in finding the bathroom in the wrong place. This indicates that our proposed action-aware navigation policy benefits from taking into account action demands in the instruction. 

\section{Conclusion}
In this paper, we take into account the instruction action demands and decompose the VLN task into a sequence of action-specific object navigation sub-tasks. To execute these sub-tasks, we further propose an action-aware navigation policy that learns different navigation abilities without requiring any manual path-instruction annotation. The proposed \sexyname achieves the best zero-shot VLN performance on two benchmark datasets (\ie, R2R-Habitat and RxR-Habitat) and outperforms the state-of-the-art supervised learning methods on RxR-Habitat.
Furthermore, our \sexyname is able to more accurately follow navigation instructions that contain specific action demands, demonstrating its potential for the scenario that needs human-robot communication and interaction.

\paragraph{Limitations and Future Works.} The proposed \sexyname achieves promising ZS-VLN performance without requiring any path-instruction annotation. 
However, it does require room region bounding box annotations for collecting action-specific datasets to train the navigators.
This may limit our method for training on the environment without the region annotation. 
To address this limitation, future work could explore leveraging well-studied scene understanding models~\cite{layout,scene_survey} to automatically infer the room region bounding boxes. By doing so, it is possible to train the action-specific navigators on a larger-scale dataset, potentially leading to improved performance.

{\small
\bibliographystyle{ieee_fullname}
\bibliography{main}
}

\clearpage

\appendix

\begin{leftline}
	{
		\LARGE{\textsc{Appendix}}
	}
\end{leftline}
\vspace{5mm}

In the supplementary, we provide more implementation details and more visualization results of our method. We organize the supplementary as follows.

\begin{itemize}[leftmargin=*]
    \item In Section~\ref{sec:supp-dataset}, we present more details on action-specific image-path dataset collection.
    \item In Section~\ref{sec:supp-navigator}, we present more details on action-specific navigator training and inference.

    \item In Section~\ref{sec:supp-prompt}, we present more details on prompt design for the instruction parser.
    \item In Section~\ref{sec:supp-baseline}, we present more details on zero-shot navigation baselines.

    \item In Section~\ref{sec:vis}, we present more visualization results.

\end{itemize}

\section{More Details on Action-Specific Image-Path Dataset Collection}
\label{sec:supp-dataset}
For learning a navigator for executing each action demand, we need to collect an action-specific image-path dataset for fine-tuning a trained ZSON model. In Section 3.3.2 of the paper, we have introduced the basic principle for collecting the episodes (\ie, the path and the corresponding goal image) in the dataset. In this section, we present more data collection details.

\begin{itemize}[leftmargin=*]
    \item \textbf{\textsc{GoPast} Dataset.} We randomly sample two points whose geometric distance is longer than 1.5m, and consider the shortest navigation path as the ground truth path. The goal image is sampled in the middle of the path facing the direction of the agent's advancement. We introduce some randomness to the angle by jittering it by $\pm45^{\circ}$.
    
    \item \textbf{\textsc{GoInto} Dataset.}
    We randomly choose a region from the scene. The start point is sampled near the entrance of this region (the geometric distance is less than 1.5m). The goal point is sampled randomly inside this region. The goal image is taken in a random direction at the goal point.

    \item \textbf{\textsc{GoThrough} Dataset.} We randomly select a region with two different entrances and sample a random point near each entrance respectively to form a path. The geometric distance from the start or end point to the entrance is less than 1.5m. Goal image is taken in the middle of the path and faces the direction of the agent's advancement.

    \item \textbf{\textsc{Exit} Dataset.} The ground truth path of the \textsc{EXIT} action is similar to the \textsc{GoInto} action besides switching the position of start point and goal point.

\end{itemize}

\begin{figure}
    \centering
    \includegraphics[width=1.0\textwidth]{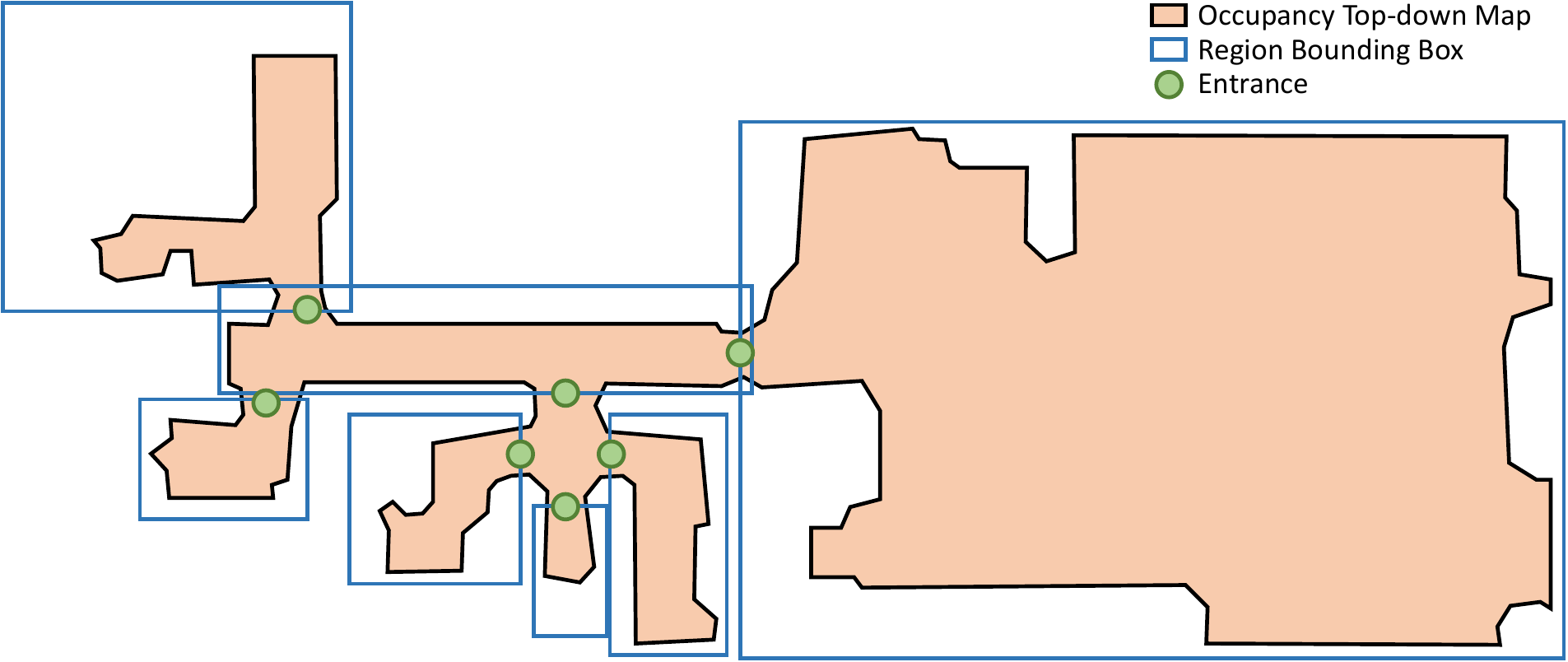}
    \caption{Obtaining entrance positions from the intersections between regions and top-down map.}
    \label{fig:entrance}
\end{figure}

We utilize room region bounding box annotations for obtaining the entrance position of regions. Specifically, we consider the intersection between the room region bounding box and the occupancy top-down map as the entrance of a region. An example is shown in Figure~\ref{fig:entrance}.
% \todo{[TODO: draw an example]}.
Since collecting \textsc{GoInto}, \textsc{GoThrough}, and \textsc{Exit} datasets requires the entrance position, we collect these three datasets from 131 scenes in HM3D~\cite{hm3d} dataset that have region bounding box annotations. For the \textsc{GoPast} dataset, we collect from all 800 scenes in the train split of HM3D. We sample 9,000 image-path pairs from each scene.

\section{More Details on Action-Specific Navigator Training and Inference}
\label{sec:supp-navigator}
We fine-tune a trained ZSON model on the action-specific dataset for learning an action-specific navigator. We use the ZSON model that is trained on agent configuration A described by Majumdar~\etal~\cite{zson}, \ie, the agent has a height of 1.25m, with a radius of 0.1m and is equipped with one 128 × 128 RGB sensor with 90$^{\circ}$ horizontal field of view. We fine-tune this model using a reinforcement learning algorithm (\ie, DD-PPO~\cite{ddppo}) for 100M steps with the same navigation reward as ZSON. This reward encourages the agent to go to the end of the path in an episode and look toward the goal image:
\begin{equation}
    r_t=r_{\rm{success}}+r_{\rm{angle-success}}-\Delta_{\rm{dtg}}-\Delta_{\rm{atg}}+r_{\rm{slack}}
\end{equation}
where $r_{\rm{success}}=5$ if \textsc{STOP} is predicted when the agent is within 1m of the goal position, $r_{\rm{angle-success}}=5$ if the agent is within 1m of the goal position and within $25^\circ$ of the goal orientation (and 0 otherwise). Besides, $\Delta_{\rm{dtg}}$ is the change in the agent's distance-to-goal, and $\Delta_{\rm{atg}}$ is the change in the agent's angle-to-goal. $\Delta_{\rm{atg}}$ is set to $0$ if the agent is not within a circle of 1m radius from the goal position. We also use a slack reward $r_{\rm{slack}}=-0.01$ to encourage the agent to reach the goal as soon as possible.

\begin{figure*}[th]
  \centering
    \includegraphics[width=1\linewidth]{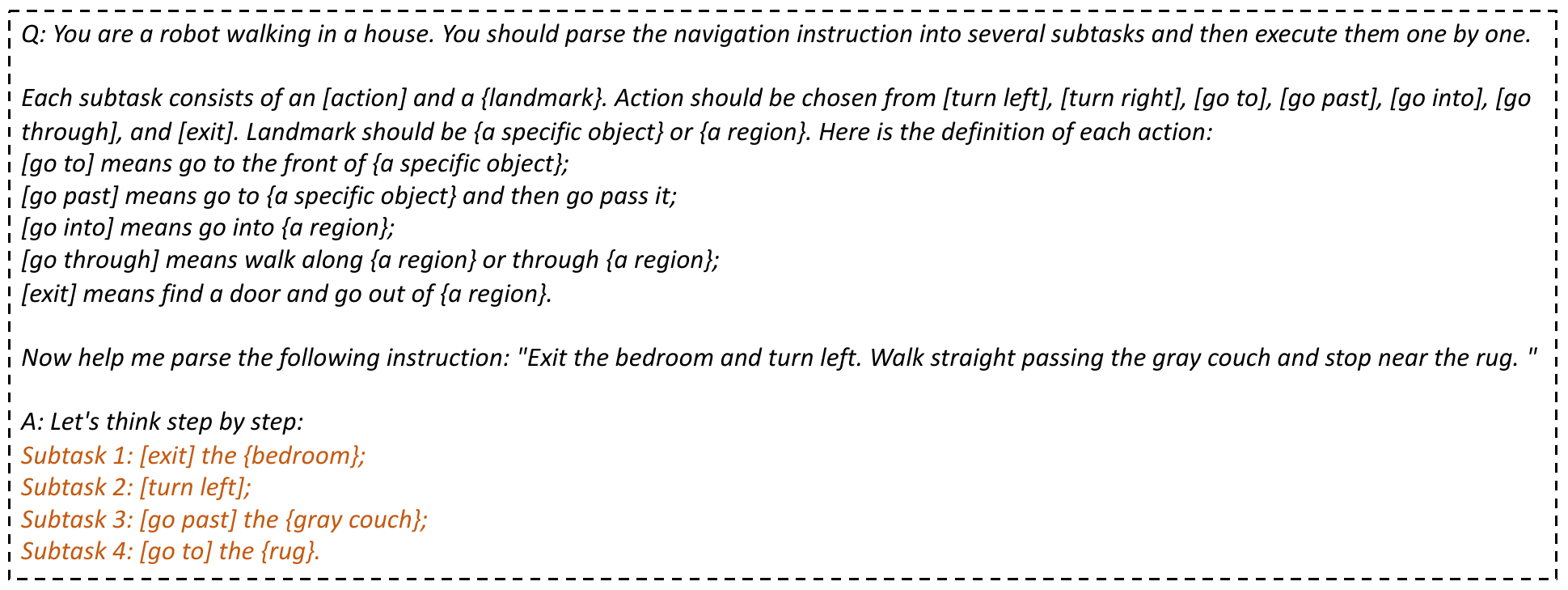}
  \caption{ An example of prompt design 1: a brief description of sub-task.
  }
  \label{fig:prompt1}
\end{figure*}

\begin{figure*}[th]
  \centering
    \includegraphics[width=1\linewidth]{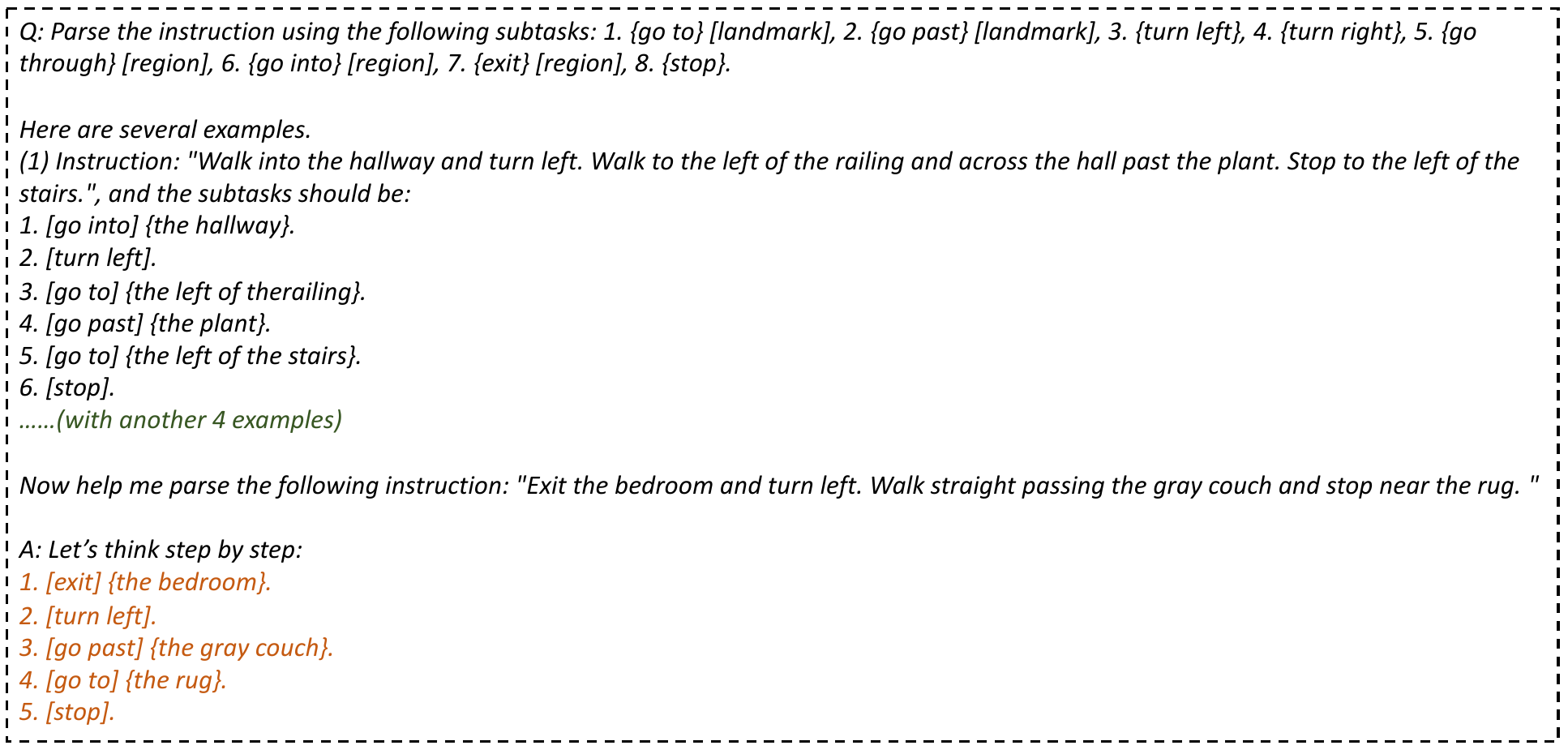}
  \caption{ An example of prompt design 2: a collection of parsing examples. This prompt design performs the best.
  }
  \label{fig:prompt2}
\end{figure*}

\begin{figure*}[th]
  \centering
    \includegraphics[width=1\linewidth]{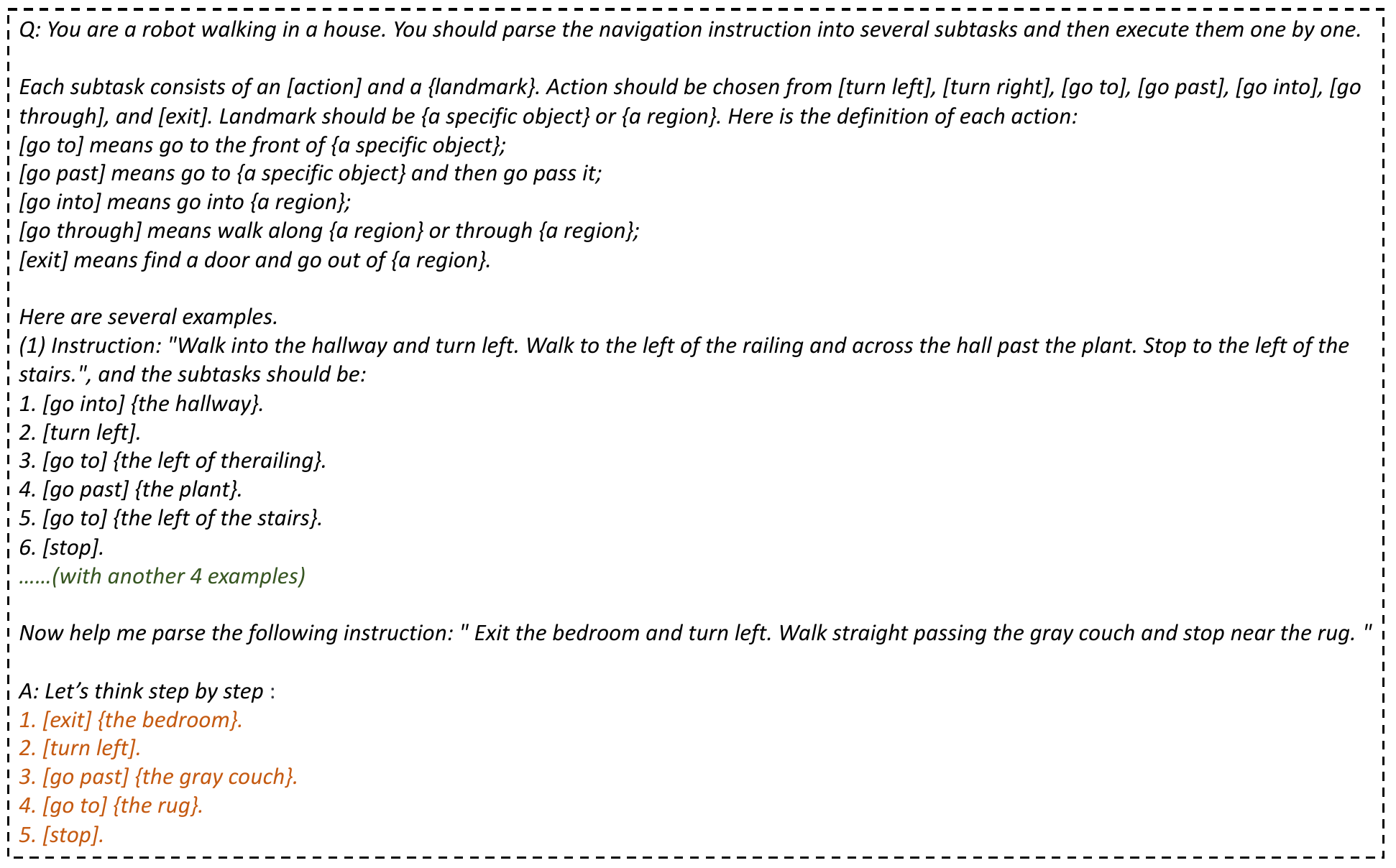}
  \caption{ An example of prompt design 3: a combination of both sub-task definition description and examples.
  }
  \label{fig:prompt3}
\end{figure*}

After fine-tuning, we use the trained navigator for executing a sub-task. Specifically, we feed the landmark (\ie, text description of an object or a region) to the CLIP for extracting goal embedding. The navigator take the current RGB observation and the goal embedding as input for predicting a low-level action for this sub-task.

\section{More Details on Prompt Design for the Instruction Parser}
\label{sec:supp-prompt}
We have tried three prompt designs for parsing instructions using the large language model GPT-3.
\begin{itemize}[leftmargin=*]
    \item \textbf{Prompt Design 1:} a brief description of each sub-task definition.
    \item \textbf{Prompt Design 2:} a collection of instruction parsing examples
    \item \textbf{Prompt Design 3:} a combination of both brief description and examples
\end{itemize}
Experimental results in Table 4 in the paper show that the second prompt design performs the best. We show the examples of these prompt designs in Figures~\ref{fig:prompt1},~\ref{fig:prompt2} and ~\ref{fig:prompt3}, respectively. We mark the GPT-3 output in \textcolor{MyBrown}{brown color}.
% \todo{[update the example figure with answer from the same long instruction]}

\section{More Details on Zero-Shot Navigation Baselines}
\label{sec:supp-baseline}
We decompose the instruction into an object navigation sub-task sequence using GPT-3 and execute these sub-tasks sequentially using four zero-shot object navigation methods.

\begin{itemize}[leftmargin=*]
\item \textbf{CLIP-Nav}~\cite{clipnav} is designed for navigating among discrete navigable nodes. The agent uses CLIP to determine which adjacent node has the highest possibility of containing landmarks and then moves to this node. To adapt it to continuous environments, we capture 4 RGB images uniformly in different directions and use CLIP~\cite{CLIP} to select one image that has the highest possibility of containing landmarks. Then, we randomly set a waypoint in that direction and use a path-planing algorithm to plan low-level actions for navigating to the waypoint. If the CLIP softmax score is higher than the threshold of 0.8, we switch to the next object navigation sub-task. For implementation convenience, we use the ``\textit{shortest\_path\_follower}'' API in the Habitat simulator for path planning, which assumes the complete occupancy top-down map is available.

\item \textbf{Seq CLIP-Nav}~\cite{clipnav} is an extended version of CLIP-Nav with an additional backtracking mechanism, which allows the agent to go back to the previous location if it cannot find the landmarks for several steps. In our implementation, we directly set the agent to the position 15 step before for performing backtracking.

\item \textbf{CoW}~\cite{cow} uses CLIP gradient for object localization and a path-planning algorithm for action determination. For implementation convenience, we use the ``\textit{shortest\_path\_follower}'' API in the Habitat simulator for path planning, which assumes the complete occupancy top-down map is available. Even using the oracle occupancy information, our \sexyname still performs better than this baseline.

\item \textbf{ZSON}~\cite{zson} uses the CLIP for encoding both image and landmark text to the same semantic feature space and then trains an image navigator for object navigation. We use the model trained on the HM3D dataset using the config A setting.
\end{itemize}

\begin{figure*}[h!]
    \centering
    \includegraphics[width=1\textwidth]{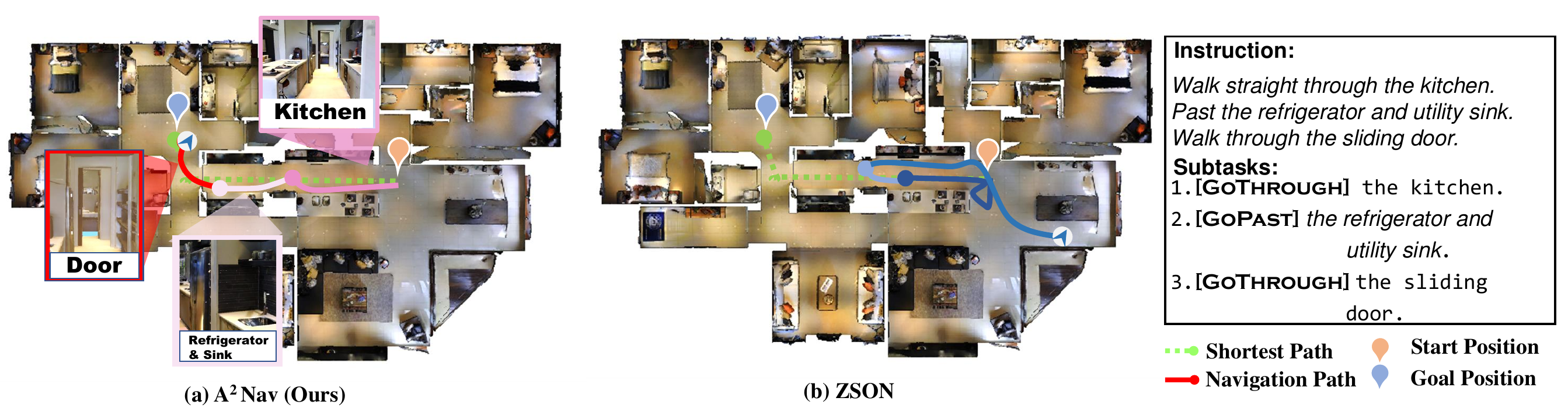}
    \caption{Visualization of the navigation path. Our method successfully goes through the kitchen, while the baseline fails to do it.}
    \label{fig:v1}
    \vspace{3mm}
\end{figure*}

\begin{figure*}[h!]
    \centering
    \includegraphics[width=1\textwidth]{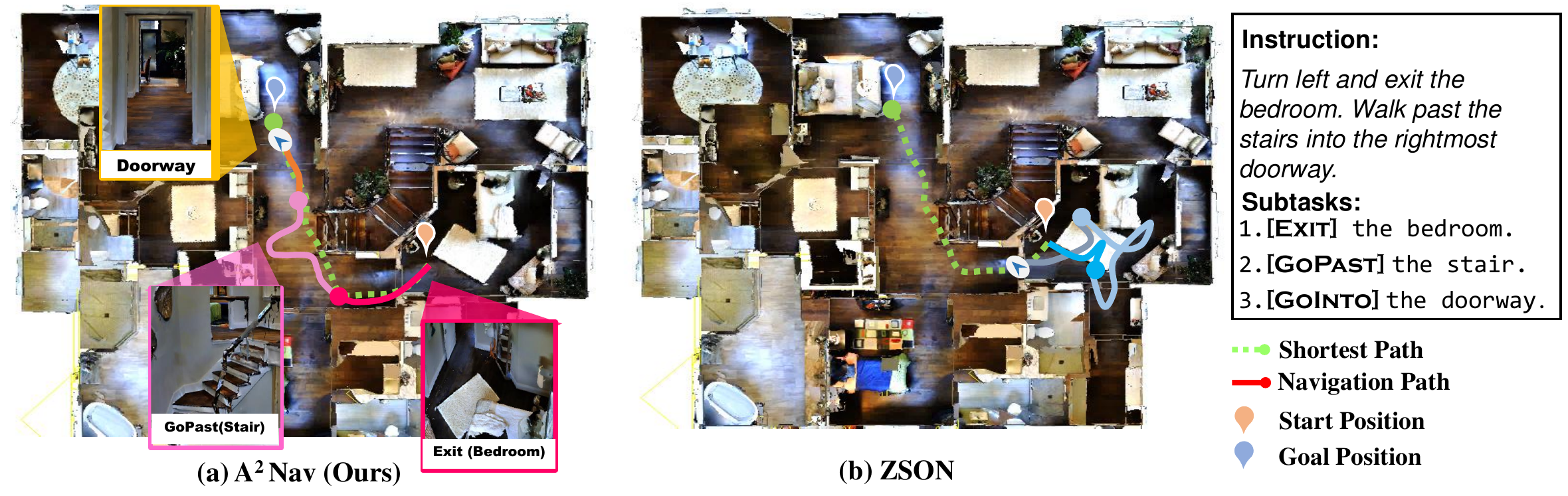}
    \caption{Visualization of the navigation path. Our method successfully exits the bedroom and goes past the stair, while the baseline is stuck in the bedroom.}
    \label{fig:v2}
    \vspace{3mm}
\end{figure*}

\begin{figure*}[h!]
    \centering
    \includegraphics[width=1\textwidth]{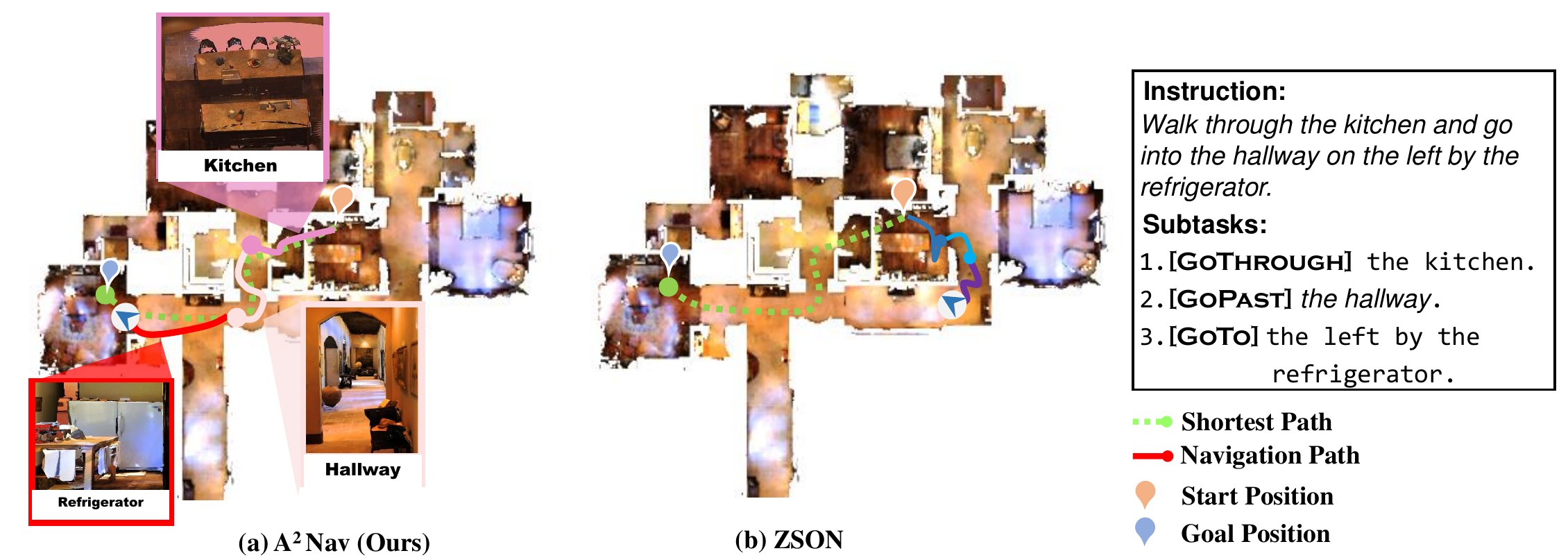}
    \caption{Visualization of the navigation path. Our method successfully goes through the kitchen and finds the refrigerator, while the baseline fails to do it.}
    \label{fig:v3}
    \vspace{3mm}
\end{figure*}

\section{More Visualization Results}
\label{sec:vis}
In this section, we provide more visualization examples for comparing the method between ZSON~\cite{zson} and ours. In Figure~\ref{fig:v1}, the instruction requires the agent to go across the kitchen area and exit this area through the door. Our \sexyname successfully follows the instruction because of the learned ``\textsc{GoThrough}'' ability, which leads the agent to completely go through the area. However, ZSON~\cite{zson} just goes to the kitchen area of a refrigerator, which directly causes the task to fail. In Figure~\ref{fig:v2} the instruction requires the agent to go past the stair which is outside the bedroom. Our \sexyname successfully exits the bedroom, navigates past the stair and then stop at the correct doorway. In contrast, the ZSON model fails to exit the bedroom and finally stop at the doorway of the bedroom incorrectly. In Figure~\ref{fig:v3}, the instruction requires the agent to get out of the kitchen and stop near the refrigerator. Our \sexyname successfully walks across the kitchen and goes by the hallway, finally finding the target. ZSON~\cite{zson} tries to go to the area which suggests the higher confidence of the kitchen, which is the opposite direction of the shortest path to the target. All examples demonstrate the effectiveness of our action-aware agent.

\end{document}

%% file: def.tex
\def\etal{{\em et al.}}

\DeclareMathAlphabet\mathbfcal{OMS}{cmsy}{b}{n}
\def\0{{\bf 0}}
\def\1{{\bf 1}}

\def\eg{\emph{e.g.}} 
\def\ie{\emph{i.e.}} 
 
 \def\vs{\emph{vs.}}
 
\def\etal{\emph{et al.}}

\usepackage{amsmath}